\documentclass[lettersize,journal]{IEEEtran}
\usepackage{amsmath,amsfonts}
\usepackage{algorithmic}
\usepackage[linesnumbered,ruled,vlined]{algorithm2e}
\usepackage{array}
\usepackage{stfloats}
\usepackage{url}
\usepackage{verbatim}
\usepackage{graphicx}
\usepackage{cite}
\usepackage{multirow}
\usepackage{multicol}
\usepackage{tikz}
\usepackage{tkz-tab}
\usepackage{pgfplots}
\usepackage{xcolor}
\usepackage{pgfplotstable}
\usepackage{balance}

\begin{document}

\title{A Global Medical Data Security and Privacy Preserving Standards Identification Framework for Electronic Healthcare Consumers}
\author{Vinaytosh Mishra, Kishu Gupta,~\IEEEmembership{Member,~IEEE,} Deepika Saxena,~\IEEEmembership{Member,~IEEE,} \\ and Ashutosh Kumar Singh,~\IEEEmembership{Senior member,~IEEE}

\thanks{Manuscript received 21 November 2023; revised 5 February 2024; accepted 2 March 2024. Date of publication 6 March 2024; date of current version 26 April 2024. This work was supported in part by the Thumbay Institute for AI in Healthcare, Gulf Medical University, Ajman, UAE, in part by National Sun Yat-sen University, Kaohsiung, Taiwan, in part by University of Aizu, Japan, and in part by Indian Institute of Information Technology Bhopal, India. (Corresponding author: Deepika Saxena.)}
\thanks{Vinaytosh Mishra is with the Thumbay Institute for AI in Healthcare, Gulf Medical University, Ajman, UAE (E-mail: dr.vinaytosh@gmu.ac.ae).}
\thanks{Kishu Gupta is with the Department of Computer Science and Engineering, National Sun Yat-sen University, Kaohsiung, Taiwan (E-mail: kishuguptares@gmail.com).}
\thanks{Deepika Saxena is with the School of Computer Science and
Engineering, The University of Aizu, Aizuwakamatsu 965-0006, Japan and also with Department of Computer Science, the University of Economics and Human Sciences, 01-043 Warsaw, Poland. (E-mail: deepika@u-aizu.ac.jp).}
\thanks{Ashutosh Kumar Singh is with the Department of Computer Science and Engineering, Indian Institute of Information Technology Bhopal, Bhopal 462003, India, and also with Department of Computer Science, the University of Economics and Human Sciences, 01-043 Warsaw, Poland. (E-mail: ashutosh@iiitbhopal.ac.in).}
\thanks{Digital Object Identifier 10.1109/TCE.2024.3373912}}

 \markboth{IEEE TRANSACTIONS ON CONSUMER ELECTRONICS,~Vol.~70, No.~1, FEBRUARY~2024}%
 {Shell \MakeLowercase{\textit{et al.}}: A Sample Article Using IEEEtran.cls for IEEE Journals}


\makeatletter
\newcommand{\removelatexerror}{\let\@latex@error\@gobble}
\def\ps@IEEEtitlepagestyle{%
	\def\@oddfoot{\mycopyrightnotice}%
	\def\@oddhead{\hbox{}\@IEEEheaderstyle\leftmark\hfil\thepage}\relax
	\def\@evenhead{\@IEEEheaderstyle\thepage\hfil\leftmark\hbox{}}\relax
	\def\@evenfoot{}%
}

\def\mycopyrightnotice{%
	\begin{minipage}{\textwidth}
		\centering \scriptsize
		1558-4127 © 2024 IEEE. Personal use is permitted, but republication/redistribution requires IEEE permission. \\
		See https://www.ieee.org/publications/rights/index.html for more information. \\This article has been accepted in IEEE Transactions on Consumer Electronics Journal © 2024 IEEE. Personal use of this material is permitted. Permission from IEEE must be obtained for all other uses, in any current or future media, including reprinting/republishing this material for advertising or promotional purposes, creating new collective works, for resale or redistribution to servers or lists, or reuse of any copyrighted component of this work in other works. This work is freely available for survey and citation.
		
	\end{minipage}
}
\makeatother

\maketitle

\begin{abstract}
Electronic Health Records (EHR) are crucial for the success of digital healthcare, with a focus on putting consumers at the center of this transformation. However, the digitalization of healthcare records brings along security and privacy risks for personal data. The major concern is that different countries have varying standards for the security and privacy of medical data. This paper proposed a novel and comprehensive framework to standardize these rules globally, bringing them together on a common platform. To support this proposal, the study reviews existing literature to understand the research interest in this issue. It also examines six key laws and standards related to security and privacy, identifying twenty concepts. The proposed framework utilized K-means clustering to categorize these concepts and identify five key factors. Finally, an Ordinal Priority Approach is applied to determine the preferred implementation of these factors in the context of EHRs. The proposed study provides a descriptive then prescriptive framework for the implementation of privacy and security in the context of electronic health records. Therefore, the findings of the proposed framework are useful for professionals and policymakers in improving the security and privacy associated with EHRs.
\end{abstract}


\section{Introduction}
\begin{figure}[!htbp]
  \centering
\begin{tikzpicture}
\begin{axis}[
width=0.4\textwidth,
height=4.0cm,
color=black,
scale only axis,
symbolic x coords={2009,2010,2011,2012,2013,2014,2015,2016,2017,2018,2019},
xtick = data,
tick label style={rotate=0},
xticklabel style = {rotate= 30,font=\scriptsize},
xlabel={\scriptsize Years},
ylabel = {\scriptsize Breaches Count},
ymin=0,
ymax=600,
ytick={0,100,200,300,400,500,600},
yticklabel style = {font=\scriptsize},
axis x line*=bottom,
axis y line*=left,
tick align = outside,
legend cell align={left},every axis x label/.style={at={(ticklabel cs: 0.5,0)},anchor=north},
legend style={draw=none,fill=none,at={(.001,.85)},anchor=west,font=\scriptsize},
]
\addplot [smooth,
color=teal,
solid,
line width=1pt,
mark size=1.5pt,
mark= star,
mark options={solid, fill=white}]
table[row sep=crcr]{
2009 16 \\
2010 149 \\
2011 138   \\
2012 145  \\
2013 152  \\
2014 150  \\
2015 105   \\
2016 77   \\
2017 71   \\
2018 55   \\
2019 54   \\
};
\addlegendentry{Loss theft of PHI and unencrypted ePHI}

\addplot [smooth,
color=magenta,
solid,
line width=1pt,
mark size=1.5pt,
mark=square,
mark options={solid, fill=white}]
table[row sep=crcr]{
2009 0 \\
2010 10 \\
2011 7   \\
2012 8  \\
2013 13  \\
2014 12  \\
2015 6   \\
2016 7   \\
2017 11   \\
2018 10   \\
2019 6   \\
};
\addlegendentry{Improper disposal of PHI/ePHI incidents }
\addplot [smooth,
color=red,
solid,
line width=1pt,
mark size=1.5pt,
mark= triangle,
mark options={solid, fill=white}]
table[row sep=crcr]{
2009 18 \\
2010 199 \\
2011 200   \\
2012 215   \\
2013 275   \\
2014 310   \\
2015 270   \\
2016 329   \\
2017 357   \\
2018 371   \\
2019 510   \\
};
\addlegendentry{Healthcare data breach}

\addplot [smooth,
color=blue,
solid,
line width=1pt,
mark size=1.5pt,
mark=diamond,
mark options={solid, fill=white}]
table[row sep=crcr]{
2009 0 \\
2010 8 \\
2011 17   \\
2012 17   \\
2013 29   \\
2014 28   \\
2015 56   \\
2016 114   \\
2017 148   \\
2018 161   \\
2019 303   \\
};
\addlegendentry{Hacking Incident}

\addplot [smooth,
color=orange,
solid,
line width=1pt,
mark size=1.5pt,
mark=otimes,
mark options={solid, fill=white}]
table[row sep=crcr]{
2009 0 \\
2010 10 \\
2011 29   \\
2012 32   \\
2013 63   \\
2014 88   \\
2015 103   \\
2016 131   \\
2017 127   \\
2018 145   \\
2019 147   \\
};
\addlegendentry{Unauthorized access/disclosure}

\end{axis}
\end{tikzpicture}%
\caption{State of Health Data Privacy and Security in the USA.} \label{fig_intro}
\end{figure}
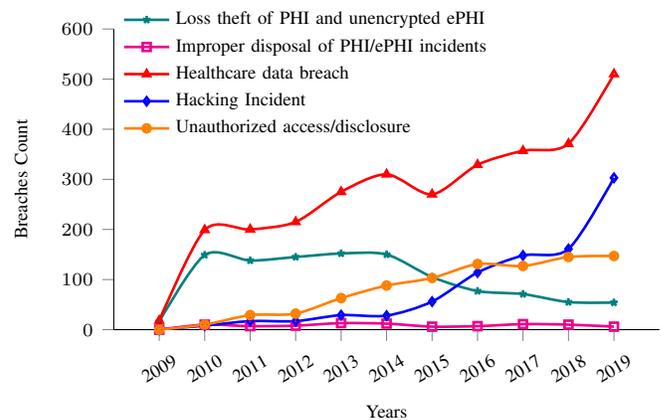

\IEEEPARstart{H}{ealthcare} consumers worldwide suffer from value crises as outcomes stagnate while care costs rise daily \cite{c1elizabeth}. Digital transformation of healthcare helps improve outcomes while simultaneously reducing cost, thus unlocking the value of the healthcare systems \cite{c2MISHRA2022100684, c31FedMUP}. Electronic Health Records (EHR) act as the backbone of digital health. The broad implementation of EHR systems enables the digital retrieval of patient records and the extraction of valuable clinical information. As a result, various additional uses have become available, including quality management, healthcare administration, and translational research. All these secondary applications aim to enhance patient care \cite{c3Friedman2013ConceptualisingAC}. The adoption of EHRs by healthcare organizations can offer numerous advantages for consumers like doctors, patients, and healthcare services. However, apprehensions regarding the privacy and security of end consumer like patient data have led to comparatively limited implementation of EMRs across various healthcare institutions \cite{c4KESHTA2021177, MAIDS}. A study in the United Arab Emirates highlights nurses' concerns about using electronic health records, including privacy, confidentiality, security, and patient safety \cite{c5Bani}. Mishra and Mishra, in their study, highlight the privacy security concerns related to EHRs \cite{c6mishra}. Fig. \ref{fig_intro} provides a glimpse of data privacy issues in the USA provided by Health Insurance Portability and Accountability Act (HIPAA) \cite{c7db}. 

As EHR systems are increasingly accessed through devices like smartphones and laptops, safeguarding sensitive patient information is crucial. This is vital for maintaining patient trust and confidentiality, as patients expect their health information to remain private. Healthcare providers must also adhere to regulations such as HIPAA, which set strict standards for handling health information. Any breach can have serious legal consequences.

Furthermore, the integrity of patient data in EHRs is critical for delivering quality healthcare. Vulnerabilities in these systems can lead to data breaches or unauthorized changes, potentially resulting in incorrect medical decisions. The rise of telemedicine has made secure remote access to EHRs more important than ever, as both providers and patients depend on consumer electronics for healthcare services. This is coupled with the need for secure data sharing and interoperability between different healthcare systems, like laboratories and pharmacies.

\par Implementing organization-wide privacy and security practices requires investments in IT infrastructure and require government subsidies for implementing such procedures \cite{c6mishra, Forecast}. Thus, they further delineated the importance of government policies in facilitating and enforcing these practices in developing countries such as India. Chenthara et al. highlight the privacy and security risks in cloud-based healthcare records and propose approaches for mitigation \cite{c8Chenthara2019SecurityAP}. For example, the pseudonymization of EHR data enables us to use the information for policy-making and clinical research without compromising the privacy and security of personal health information (PHI). It involves substituting identification data with pseudonyms, which serve as subject identifiers. This method can address privacy, confidentiality, and data integrity concerns.
\par Pseudonyms also have a role in facilitating linkability \cite{c9article}. Numerous recent studies cite blockchain as one of the potential technologies for addressing the privacy and security concerns related to EHRs \cite{c10article, c119899708, hac1-10409548}. Tang et al., in their paper, discuss an efficient scheme \cite{c12Tang2019AnEA}, while Guo et al. propose a secure attribute-based signature scheme with multiple authorities for blockchain in EHRs \cite{c13267415}. Al Mamun et al., in their paper, provide a comprehensive review of the use of blockchain in EHRs and suggest future research directions \cite{c14article}. But apart from technical factors, behaviour and awareness also play an important role in managing the privacy and security of personal health information \cite{c15article, hac2-10409548}. Achieving privacy and security requires overhauling infrastructure, processes and practices in a healthcare organization and requires resources to implement it effectively \cite{infocomp}.
\par Similarly, ensuring secure health information exchange across different organizations requires standardization of security measures, such as global definitions of professional roles, international standards for patient consent and semantic interoperable audit logs \cite{c16article}. Organizations should consider the compromise of PHI as a quality defect and use appropriate standards as quality management systems to minimize it. Next, this study reviews six such standards: Lei Geral de Proteção de Dados in Portuguese (LGPD), the Digital Personal Data Protection Act (DPDPA), the Health Insurance Portability and Accountability Act (HIPAA), the National Institute of Standards and Technology (NIST), General Data Protection Regulation (GDPR), and Office of the National Coordinator for Health Information Technology (ONC). 

\begin{figure}[!htbp]
\centering
\includegraphics[width=0.49\textwidth]{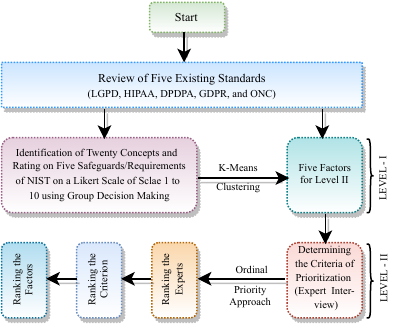}
\caption{Workflow for two-level approach framework.}
\label{fig_two-level}
\end{figure}

Although there are plenty of studies on privacy and security issues related to personal health information, there is a lack of studies reviewing, categorization, and prioritization of best practices and this article attempts to fill this research gap. In this context, a novel \textbf{G}lobal Medical \textbf{D}ata \textbf{S}ecurity and \textbf{P}rivacy Preserving \textbf{S}tandard Identification for Electronic healthcare Consumers \textbf{(GDSPS)} is proposed. The workflow for two-level approach used in this framework is summarized in Fig. \ref{fig_two-level}.
\subsection{Key contributions}
The key contributions of this paper are listed below:
\begin{enumerate}
\item{A novel GDSPS framework is proposed to provide a worldwide appropriate and homogeneous medical benchmark standard to furnish digital healthcare efficiently with enhanced privacy and security over existing diverse medical standards.}
\item{Categorization and assessment of identified key concepts in homogeneous categories to facilitate effective decision-making.}
\item{Prioritisation of identified categories on three criteria Ease, Efficacy and Cost.}
\item{Designed a descriptive and prescriptive framework for implementing privacy and security in the context of EHRs.}
\end{enumerate}
\textit{Paper outline:} The remaining paper is structured as follows. The next section \ref{secstd} provides an academic review of six standards related to privacy and security of EHRs, followed by proposed GDSPS framework in section \ref{secproposed} along-with operational design. Thereafter, performance evaluation of GDSPS framework explaining experimental setup, dataset, results and a brief discussion is described in section \ref{secper}. Section \ref{seccon}, finally concludes with a summary of the study and its implications for theory and practice.
\section{Key Security Standards} \label{secstd}
\begin{figure}[!ht]
\centering
\includegraphics[width=\linewidth]{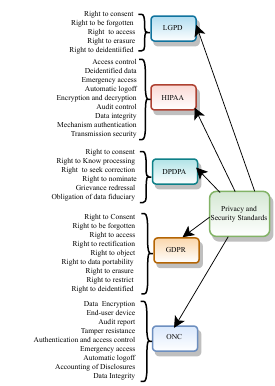}
\caption{Key security standards and selected concepts.}
\label{fig_std}
\end{figure}

This section provides the details of key data privacy and security standards in healthcare field. The visual representation of six standards and associated concepts is depicted in Fig. \ref{fig_std}. 
\begin{enumerate}
    \item {\textit{LGPD} is a Portuguese term referring to the comprehensive legislation governing data protection and privacy in Brazil. This legal framework was enacted to streamline the regulation of personal data processing and list five rights of patient health information as (1) Right to consent, (2) Right to be forgotten, (3) Right to Access, (4) Right to erasure, and (5) Right to de-identified \cite{c17article, c18article}.}  
    \item {\textit{GDPR} is a comprehensive data protection and privacy regulation enacted by the European Union (EU) to ensure the privacy and security of healthcare  personal data of EU citizens. The salient point of the GDPR is (1) Right to consent and (2) Right to be forgotten, (3) Right to access, (4) Right to rectification, (5) Right to object, (6) Right to data portability, (7) Right to erasure, (8) Right to Access Control, (9) Right to de-identified \cite{c17article, c18article}.} 
    \item {\textit{HIPAA} is a US federal law enacted in 1996. It is the most widely used privacy and security standard and has been adopted by many countries. The salient points considered in HIPAA are guidelines for (1) Access control, (2) de-identified data, (3) Emergency access procedure, (4) Automatic logoff, (5) Encryption and decryption, and (6) Audit Controls. (7) Data integrity, (8) Mechanism authentication, and (9) Transmission security \cite{c17article, c19article}.} 
    \item {Amid rapid digitalization in healthcare, India passed the \textit{DPDPA} in August 2023. The act provides guidelines for the right to protection of personal data and provision to process sensitive personal data for lawful purposes. The salient points of the DPDPA are (1) Right to consent, (2) Right to know about processing, (3) Right to seek correction, (4) Right to nominate another person to exercise rights, (5) Right for grievance redressal, and (6) Obligation of data fiduciary \cite{c21moe}.} 
    \item {\textit{ONC} plays a pivotal role in advancing the adoption and meaningful use of Electronic Health Records (EHRs) in the United States healthcare system. It creates guidelines and procedures for (1) Data encryption, (2) End user device, (3) Audit report, (4) Tamper resistance, (5) Authentication and access control, (6) Emergency access, (7) Automatic logoff, (8) Optional accounting of disclosures, and (9) Data integrity \cite{c17article, c22barker}.} 
    \item {\textit{NIST} in the USA is actively involved in technological and standards development regarding Electronic Health Records (EHRs). NIST provides guidelines and best practices for maintaining EHRs. Five key points related to NIST guidelines are (1) Administrative safeguards (AS), (2) Physical safeguards (PS), (3) Technical safeguards (TS), (4) Organizational requirements (OR), and (5) Policy and procedures requirements (PPR) \cite{c17article, c20article}.} 
\end{enumerate}  
\section{GDSPS Framework} \label{secproposed}
\begin{figure}[!ht]
\centering
\includegraphics[width=\linewidth]{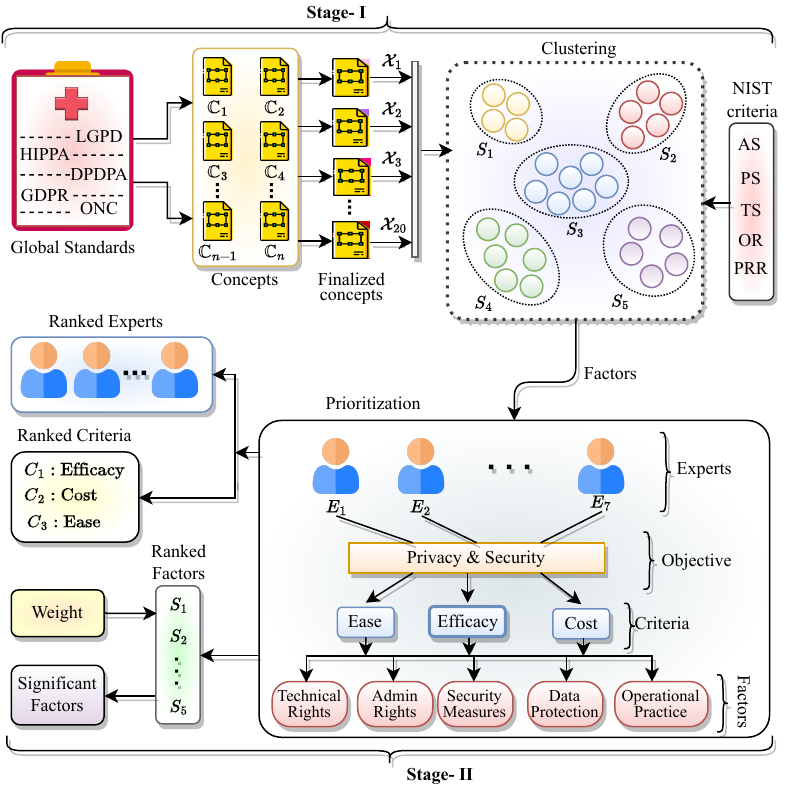}
\caption{Framework of proposed GDSPS.}
\label{fig_proposed}
\end{figure}

The proposed GDSPS framework considers the provisions of the five global medical standards: \{LGPD, GDPR, HIPAA, DPDPA, ONC\}. Various standards as discussed in section \ref{secstd} involve numerous concepts: $\mathbb{C} = \{\mathbb{C}_1, \mathbb{C}_2, ..., \mathbb{C}_n\}$. The Framework of the proposed GDSPS model is depicted in Fig. \ref{fig_proposed}. A comprehensive study after necessary filtration and removing the redundancy among these diversed medical concepts ($\mathbb{C}$), generates a virtuous list of twenty prominent concepts: $\mathcal{X} = \{\mathcal{X}_1, \mathcal{X}_2, ..., \mathcal{X}_{20}\}$ described as follows: (1) Right to consent, (2) Right to access, (3) Right to erasure, (4) Right to de-identified, (5) Right to rectification, (6) Right to object, (7) Right to data portability, (8) Right to access control, (9) Automatic logoff, (10) Encryption and decryption, (11) Audit controls, (12) Data Integrity, (13) Mechanism authentication, (14) Transmission security, (15) Right to nominate, (16) Grievance redressal, (17) End user device, (18) Tamper Resistance, (19) Emergency Access, and (20) Accounting of disclosures. Table \ref{table:1} brings forward a brief description of the prominent twenty concepts.
\par GDSPS uses K-means clustering for partitioning concepts into $k$ clusters in which each observation belongs to the cluster with the nearest mean \cite{c239072123}. This approach helps to reduce identified concepts into homogeneous and manageable groups for further prioritization \cite{c24mishra}. The number of clusters was kept at five, adhering to the golden rule of 7 (plus minus two), also known as \textit{Miller's Law} \cite{c25}.

The study involved focus group of seven health information technology experts: $E= \{ E_1, E_2, ..., E_7\}$ based in India, the USA, and the UAE. Among them, four are employed in companies primarily focused on information technology, while the remaining three work in hospital settings. Each of these experts possesses a minimum of five years of experience in the field of health informatics. These $E$ rate the twenty key concepts, on five NIST criteria: \{AS, PS, TS, OR, PRR\} elaborated in section \ref{secstd}. The rating of the factor was done on a \textit{Likert scale} of one to ten, where one is least favourable (not at all related) while ten is most favourable (extremely related). For dimensionality reduction, K-means clustering method was employed. 
\begin{table*}[!htbp] 
	\centering
    \caption{Description of Key Concepts}
	\label{table:1}
		\begin{tabular}{|p{3cm}|c|p{12.5cm}|}
		\hline
        \textbf{Key Concept} & \textbf{Notation} & \textbf{Concept Description} \\ \hline 		
         Right to consent & CNCPT1 & The right of patients to provide informed consent before their health information is collected \\ \hline 
         Right to access & CNCPT2 & The right of patients to access and view their health information stored in an EHR  \\ \hline 
         Right to erasure & CNCPT3 & A patient can request removal of their data from an EHR system, subject to legal and regulatory constraints \\ \hline 
         Right to deidentified & CNCPT4 & The right to request removal of personally identifiable information before using data for research \\ \hline 
         Right to rectification & CNCPT5 & Patients can request corrections or updates to their health information within an EHR to ensure accuracy \\ \hline
         Right to object & CNCPT6 & Patients can object to certain uses of their health information within an EHR, such as for research purposes, based on personal preferences or beliefs \\ \hline 
         Right to data portability & CNCPT7 & Patients can request their health information in a portable format, enabling them to share it  with other healthcare providers or systems  \\ \hline 
         Right to access control & CNCPT8 & Hospitals must implement access controls to ensure that only authorized personnel can view or modify patient data within an EHR  \\ \hline 
         Automatic logoff & CNCPT9 & EHR systems automatically log users out after inactivity to prevent view or modify patient data unauthorized access to patient data  \\ \hline 
         Encryption and decryption & CNCPT10 & EHR systems use encryption to secure patient data both during transmission and while stored Decryption is process of converting encrypted data back to its original form \\ \hline 
         Audit controls & CNCPT11 & EHRs maintain audit logs that record who accessed patient data, what actions were taken, and when it was taken \\ \hline 
         Data integrity& CNCPT12 & Ensuring patient data stored within an EHR remains accurate, complete, and unaltered \\ \hline 
         Mechanism authentication & CNCPT13 & The process of verifying the identity of users before granting them access to the EHR system using authentication mechanisms like passwords, bio-metrics/smart cards \\ \hline
         Transmission security& CNCPT14 & Implementing safeguards to protect patient data between different systems or devices \\ \hline
         Right to nominate & CNCPT15 & Patients can nominate individuals or caregivers who can access their health information \\ \hline
         Grievance redressal & CNCPT16 & Establishing processes for patients to address concerns or complaints about their EHR data privacy or security \\ \hline
         End-user device& CNCPT17 & The devices healthcare professionals use to access and interact with the EHR system should not be compromised \\ \hline
         Tamper resistance & CNCPT18 & Designing EHR systems and devices to resist unauthorized changes or tampering to ensure data integrity \\ \hline
         Emergency access& CNCPT19 & Granting authorized personnel temporary access to EHRs in emergencies to provide timely patient care \\ \hline
         Accounting of disclosures & CNCPT20 & Maintaining records of who accessed and shared patient information within an EHR, ensuring transparency and accountability \\ \hline
			\noalign{\smallskip}
		\end{tabular} 
\end{table*}
\subsection{Data Clustering}

The twenty key concepts ($\mathcal{X}_1, \mathcal{X}_2, ..., \mathcal{X}_{20}$) where each concept is a 5-dimensional real vector, k-means clustering in this study aims to partition these twenty concepts into five sets $\mathbb{S}= \{\mathbb{S}_1, \mathbb{S}_2, \mathbb{S}_3, \mathbb{S}_4, \mathbb{S}_5\}$ such that the within-cluster sum of squares (WCSS) is minimized. The objective of the k-means clustering for this study is given in Eqs. (\ref{eq1}) and (\ref{eq2}). 
\begin{gather}
\label{eq1} 
    \underset{\mathbb{S}}{\underbrace{\mathrm{argmin}}}=  \sum_{i=1}^5\sum_{\mathcal{X} \in \mathbb{S}_i}||\mathcal{X}-\mu_i||^2 \\
    \label{eq2} 
     \underset{\mathbb{S}}{\underbrace{\mathrm{argmin}}}= \sum_{i=1}^5|\mathbb{S}_i|Var \mathbb{S}_i
\end{gather}
where $\mu_i$ is the \textit{mean of points} in  given by Eq. (\ref{eq3}).
\begin{equation}
\label{eq3}
    \mu_i = \frac{1}{|\mathbb{S}_i|}\sum_{\mathcal{X} \in \mathbb{S}_i}\mathcal{X}
\end{equation}
here, $|\mathbb{S}_i|$ is the size of the set, $\mathbb{S}_i$ and $||\,||$ is the usual $L^2$ norm. The clusters are now used as factors affecting the implementation of privacy and security in EHRs.
\subsection{Critical Parameters Prioritization}
The problem hierarchy mentioned in the description is depicted in the Fig. \ref{fig_proposed} as Stage 2. The OPA is need for prioritisation of the clustered key concept. The high priority factors need to be addressed first in case of resource constrained. For prioritizing key factors ordinal priority approach (OPA) having three levels of hierarchy is utilized. The top level consists of objectives. While, the second level consists of criteria of evaluation. The third and final level consist of factors that need to be prioritized. The three criteria for prioritization were ease, efficacy, and cost associated with implementation. The seven experts used earlier in the study were surveyed to solve the prioritization problem: Max $Z$ using Eqs. (\ref{eq4}-\ref{eq6}) here, Z is unrestricted in sign and it denotes the multi objective function of nonlinear formulation calculating the rank of a factor on basis of the NIST criteria of evaluation.More information about the approach can is discussed in detail in seminal paper by \cite{ATAEI2020105893}.

\begin{gather}
    \label{eq4}
    Z\leq r_i\,(r_j\,(r_k\,(W_{ijk}^{r_k} - W_{ijk}^{r_{k+1}}))) \quad \forall \, i,\, j, \,\textrm{and} \, r_m \\ 
    \label{eq5}
    Z \,\leq \, r_i\, r_j\, r_m\, W_{ijk}^{r_m} \quad \forall \, i,\,j \, \textrm{and} \, r_k \\
    \label{eq6}
    \sum_{i=1}^p \sum_{j=1}^n \sum_{k=1}^m W_{ijk} = 1; \quad W_{ijk} \geq \, 0 \quad  \forall \, i, \, j \, \textrm{and} \, k 
\end{gather}

In the model $r_i\,(i=1,2,...,p)$ represents the rank of the expert $i$, $r_j\,(i=1,2,...,n)$ represents the rank of the criteria $j$ and $r_k\,(i=1,2,...,m)$ represents the rank of the factor $k$. Furthermore, $W_{ijk}$ represents the weight of the factor $k$ on criterion $j$ by expert $i$. After solving the linear programming formulation of OPA, the \textit{weight of each factor} is given by the Eq. (\ref{eq7}).
\begin{equation}
\label{eq7}
    W_k = \sum_{i=1}^p \sum_{j=1}^n W_{ijk} \quad \forall k
\end{equation}
Similarly, the \textit{weight of each criterion} is given by the Eq. (\ref{eq8}).
\begin{equation}
\label{eq8}
    W_j = \sum_{i=1}^p \sum_{k=1}^m W_{ijk} \quad \forall j
\end{equation}
Finally, the \textit{weight of each expert} is given by the Eq. (\ref{eq9}).
\begin{equation}
\label{eq9}
    W_i = \sum_{j=1}^n \sum_{k=1}^m W_{ijk} \quad \forall i
\end{equation}

The details of the mathematical formulation can be found on the website created by researchers who developed OPA \cite{ATAEI2020105893}. They also provide open-source software for solving this OPA formulation and assessing the reliability of results. 

Experts rated twenty key concepts on three criteria: ease, efficacy, and cost, on one to ten scale. The average of the responses was taken as input for the K-means clustering problem. Once factors were ascertained, OPA was employed for decision-making. Each expert ranked criteria first, followed by ranking each factor for each criterion. Weight of each expert's decision was taken as equal. Operational summary of the proposed work is depicted in Algorithm \ref{algo_ehc}.
	\begin{algorithm}[!htbp]
		\caption{GDSPS: Operational Summary}
		\label{algo_ehc}
		Input: Key Concept Set $\mathbb{C}= \{\mathbb{C}_1, \mathbb{C}_2, ..., \mathbb{C}_n\}$,  NIST criteria \{AS, PS, TS, OR, PRR\} \\
		Output: Clusters, $\mathbb{S}= \{\mathbb{S}_1, \mathbb{S}_2,\mathbb{S}_3, \mathbb{S}_4, \mathbb{S}_5\}$, ranked factor $W_k$, ranked criteria $W_j$, ranked expert $W_i$ \\
		\textbf{repeat} \\
        Initialize centroids $\{\mu_1, \mu_2, ..., \mu_5\}$, randomly from $\mathbb{C}$ \\
        \For {each key concept n = (1, ..., N)}{
        Partition sample $\mathcal{X}_n$ by $\lambda_n = \underset{\mathbb{S}}{\underbrace{\mathrm{argmin}}}= \sum_{i=1}^5\sum_{\mathcal{X} \in \mathbb{S}_i}||\mathcal{X}-\mu_i||^2$  and 
        append to corresponding cluster $\mathbb{S}_{\lambda_n} = \mathbb{S}_{\lambda_n} \cup \{\mathcal{X}_n\}$ \\
        }
	 	\For {each cluster i = (1, 2, ..., 5)}{
	 	Update new centroids $\mu_i = \frac{1}{|\mathbb{S}_i|}\sum_{\mathcal{X} \in \mathbb{S}_i}\mathcal{X}, i \in \{1,2,...,5\} $ \\
	   }
        \textbf{untill} no changes in all centroids \\
        Prioritizing experts, criteria and factors using Eqs. (\ref{eq4}-\ref{eq9}) to impart more security to EHR \\
	\end{algorithm}
\section{Performance Evaluation} \label{secper}
\subsection{Experimental Setup}
The experiment and simulation was carried out on the machine having 11th Gen Intel\textsuperscript{\textregistered} Core \textsuperscript{TM} i7-1195G7 CPU, with a clock speed of 2.92 GHz. This computational system utilizes a 64-bit Windows 11 Home 22H2 version Operating system. The  system is using RAM capacity 32.0 GB (31.8 GB usable). The software used for the K-means clustering problem is IBM SPSS 26, while for solving linear programming formulation of ordinal priority approach is the web version of OPA Solver 1.4. Seven experts were asked to rate twenty concepts on five NIST criteria. The average response value is used to form a clustering problem response matrix. The data is collected on an ordinal scale of one to ten, with one being minimum and ten being the maximum. None of the data was found incomplete. In this study, the number of clusters was pre-decided, and principal component analysis was used for dimensionality reduction. The distance metric used for the study is Euclidean distance, and convergence was achieved after three iterations. The OPA problem's objective function and constraints are explained in detail in section \ref{secproposed}.
\subsection{Dataset}
The data set used in the study is collected through a survey of focused groups. The experts for the focus group were selected using a purposive sampling approach. The responses were collected through a structured instrument. Questionnaire was explained to the respondents and were assured that the responses were complete. Unidentified data can be provided on request.Seven experts ranked the twenty concepts on five NIST criteria on a scale of one to ten for clustering problem to get five factors. While these five factors were ranked by seven experts on basis of three criteria ease, efficiency, and cost. The deidentified data set is uploaded in online repository and can be assessed at \cite{vinaydata}.
\subsection{Results}
The descriptive statistics for the K-means problem are depicted in Table \ref{table:2}. ADMIN, PHY, and TECH represent Administrative, Physical and Technical Safeguards, while ORG and PLCY represent organizational, policy and procedure requirements.
\begin{table}[!htbp] 
	\centering
    \caption{Descriptive statistics for clustering key concepts}
	\label{table:2}
	\resizebox{0.9\columnwidth}{!}{
        \tiny
		\begin{tabular}{|c||c|c|c|c|c|}
		\hline
        \textbf{Criterion} & \textbf{N} & \textbf{Min} & \textbf{Max} & \textbf{Mean} & \textbf{Std Deviation} \\ 
         \hline 		
        ADMIN & 20 & 4 & 10 &7.20 & 2.285 \\ \hline
        PHY & 20 & 3 & 9 & 6.15 & 2.183  \\ \hline
		TECH & 20 & 8 & 10 & 9.05 & 0.887 \\ \hline
        ORG & 20 & 4 & 8 & 5.75 & 1.372 \\ \hline
        PLCY & 20 & 8 & 9 & 8.35 &0.489 \\ \hline 
			\noalign{\smallskip}
		\end{tabular}}
\end{table}

Convergence was achieved in the third iteration due to a small change in cluster centres. The minimum distance between initial centres is 1.732. The distance between the final cluster centres is given in Table \ref{table:3}. The results show that the centres of the clusters are distinctly separated from each other. 
\begin{table}[!htbp] 
	\centering
    \caption{Distances between Final Cluster Centers}
	\label{table:3}
	\resizebox{0.9\columnwidth}{!}{
        \tiny
		\begin{tabular}{|c|c|c|c|c|c|}
		\hline
        \textbf{Cluster} & \textbf{1} & \textbf{2} & \textbf{3} & \textbf{4} & \textbf{5} \\ 
         \hline 		
        1 & 0 & 1.744 & 5.970 & 7.046 & 5.466 \\ \hline
        2 & 1.744 & 0 & 7.681 & 8.660 & 6.931  \\ \hline
		3 & 5.970 & 7.681 & 0 & 2.000 & 3.262 \\ \hline
        4 & 7.046 & 8.660 & 2.000 & 0 & 2.728 \\ \hline
        5 & 5.466 & 6.931 & 3.262 & 2.728 & 0 \\ \hline 
			\noalign{\smallskip}
		\end{tabular}}
\end{table}

Next, we discuss the Analysis of Variance (ANOVA) table for the K-means cluster analysis in Table \ref{table:4}. ANOVA table is used to assess the significance of differences between clusters formed during the clustering process. A p-value \textgreater 0.05 suggests that the difference is significant at a confidence interval of 95\%.
\begin{table}[!ht]
\caption{ANOVA Table for Clustering\label{table:4}}
\centering
\resizebox{0.9\columnwidth}{!}{
\tiny
\begin{tabular}{|c|c|c|c|c|c|c|}
\hline
\multirow{2}{*} & \multicolumn{2}{c|}{\textbf{Cluster}} & \multicolumn{2}{c|}{\textbf{Error}} & \multirow{2}{*}{$\mathcal{F}$} & \multirow{2}{*}{$\delta$} \\ \cline{2-3} \cline{4-5} 
& $\varepsilon$ & $\pi$ & $\varepsilon$ & $\pi$ & & \\
\hline
        ADMIN & 24.800 & 4 & 0.000 & 15 & - & - \\ \hline
        PHY & 22.438 & 4 & 0.053 & 15 & 420.7 & 0.001 \\ \hline
		TECH & 3.438 & 4 & 0.080 & 15 & 42.96 & 0.000 \\ \hline
        ORG & 8.738 & 4 & 0.053 & 15 & 163.8 & 0.001 \\ \hline
        PLCY & 0.938 & 4 & 0.053 & 15 & 17.58 & 0.000 \\ \hline 
\end{tabular}}
$\varepsilon$: Mean square error, $\pi$: Degree of freedom , $\mathcal{F}$: F-Statistics,\\ $\delta$: Significance.
\end{table}

Finally, we analyzed the clusters and gave them descriptive names based on the concepts associated with them. Five identified clusters and their details are provided in Table \ref{table:5}.
\begin{table}[!ht]
\caption{Factor Clustering\label{table:5}}
\centering
\resizebox{0.9\columnwidth}{!}{
\begin{tabular}{|c|p{3.7cm}|c|}
\hline
\textbf{Factors} & \textbf{Key Concepts} & \textbf{Notation} \\ \hline 
\multirow{5}{*}{\shortstack{Technical \\Rights}} & Right to access & CNCPT2   \\ 
& Right to earsure & CNCPT3   \\ 
& Right to deidentified & CNCPT4   \\
& Right to rectification & CNCPT5    \\
&Right to data portability & CNCPT7   \\ \hline \hline
\multirow{4}{*}{\shortstack{Admin\\ Rights}} & Right to consent & CNCPT1  \\
& Right to object & CNCPT6  \\
& Right to access control & CNCPT8  \\
& Right to nominate & CNCPT15 \\ \hline \hline
\multirow{5}{*}{\shortstack{Security\\ Measures}} & Automatic logoff & CNCPT9  \\
&Encryption and decryption & CNCPT10  \\
&Tamper resistance & CNCPT18  \\
&Mechanism Authentication & CNCPT13  \\
&Transmission Security & CNCPT14  \\ \hline \hline
\multirow{3}{*}{\shortstack{Data \\Protection}} & Audit control & CNCPT11  \\
&Data Integrity & CNCPT12 \\
&Accounting of disclosures & CNCPT20  \\ \hline \hline
\multirow{3}{*}{\shortstack{Operational\\ Practices}} & Grievance redressal & CNCPT16   \\
&End user device & CNCPT17  \\
&Emergency access & CNCPT19 \\ \hline
\end{tabular}}
\end{table}

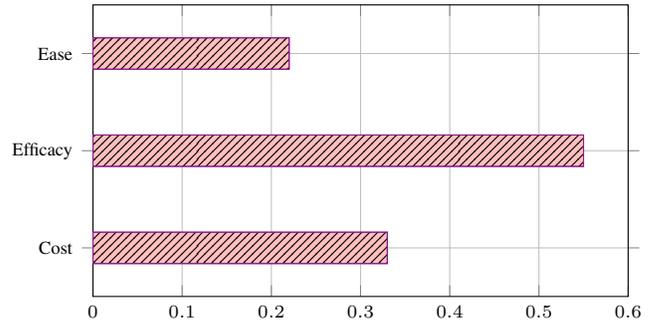
\begin{figure}[!ht]
    \centering
    \begin{tikzpicture}[node distance = 1cm,auto,scale=.99, transform shape]
        \begin{axis} [
        height=5.5cm, xmin=0, xmax=0.6, xbar,ytick={0,1,2},yticklabels={Cost,Efficacy,Ease}, grid=major,
        bar width=12,ymin=-0.5,ymax=2.5, width=0.99\columnwidth,tick label style={font=\scriptsize},]
        \addplot+[mark options={fill=pink},fill=pink, draw=violet!90, postaction={pattern=north east lines}]
        coordinates{
        (0.33,0) 
        (0.55,1) 
        (0.22,2) 
        };
        \end{axis}
    \end{tikzpicture}
    \caption{Importance of Criteria of evaluation of Factors.}
    \label{fig_criteria}
\end{figure}

Now, we will discuss the results of the OPA approach. The weight of each expert in the study is the same at 0.14. The priority weight of criteria, namely Ease, Efficacy, and Cost, was found to be 0.22, 0.55, and 0.33, respectively. The result suggests that the efficacy of the privacy and security measures is the most important criterion for evaluating factors associated with EHRs Fig. \ref{fig_criteria}. 

\begin{figure}[!ht]
    \centering
\begin{tikzpicture}[node distance = 1cm,auto,scale=.99, transform shape]
\begin{axis}[
        ymin=0, ymax=0.4,x tick label style={rotate=0,anchor=center},
        ytick={0,0.1,0.2,0.3,0.4},
        height=5.5cm,
         ybar,width=0.49\textwidth,grid=major,
        ylabel={\scriptsize Importance},
        bar width=12,
        legend style={at={(0.5,-0.18)},
        anchor=north,legend columns=-1},
        symbolic x coords={TR,AR,SM,DP,OP},
        xtick=data,tick label style={font=\scriptsize},
        nodes near coords align={vertical},
    ]
\addplot+[draw=orange!90, postaction={pattern= horizontal lines}]
coordinates {(TR,0.34) (AR,0.18) (SM,0.08)(DP,0.08) (OP,0.32)};
\end{axis}
\end{tikzpicture} \\
\scriptsize TR: Technical Right, AR: Admin Right, SM: Security Measures, \\DP: Data Protection, OP: Operational Practices.
\caption{Importance of Factors for privacy and security.} \label{fig_factors}
\end{figure}
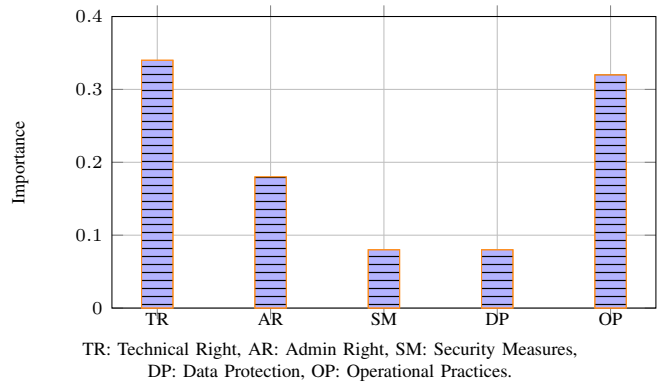

The priority weights of Technical Rights, Admin Rights, Security Measures, Data Protection, and Operational Practices were found to be 0.34, 0.18, 0.08, 0.08, and 0.32. The results suggest that Technical Rights are the most important factor, followed by Operational Practices and Admin Rights. Strangely, data protection and security measures got the lowest ranking of the lot. Priority weight for the three criteria and five factors is depicted in Fig. \ref{fig_factors}. 

Next, \textit{Kendall's W} (a coefficient of concordance) was used to assess the degree of agreement among multiple decision-makers when ranking items. Kendall's W for criteria was found to be Ease (0.92), Efficacy (0.90), and Cost (0.93), while Kendall's W among criteria was found to be 0.76. The global confidence level was 0.99, suggesting the model suits sensitive problems. Next, the uncertainty analysis of the input of decision-makers was performed. Eight hypotheses were tested on whether the length exceeds the threshold value of all criteria and factors that fulfil the threshold test. The results of the uncertainty analysis are given in Table \ref{table:6}.
\begin{table}[!htbp]
\caption{Uncertainty analysis of the decisions} \label{table:6}
\centering
\resizebox{0.99\linewidth}{!}{
\renewcommand*{\arraystretch}{1.25}
    \begin{tabular}{|c|c|c|c||c|c|c|c|c|} 
  \hline
  \multirow{2}{*}{\textbf{Analysis}} & \multicolumn{3}{c||}{\textbf{Criteria}} & \multicolumn{5}{c|}{\textbf{Factors}} \\ \cline{2-4} \cline{5-9} 
        & $\textbf{C1}$ & $\textbf{C2}$ &$ \textbf{C3}$ &$\textbf{F1}$ &$\textbf{F2}$ &$\textbf{F3}$ &$\textbf{F4}$ &$\textbf{F5}$ \\
\hline
    $\underline{\omega}$& 0.18 & 0.55 &0.18 &0.24 &0.16 &0.05 &0.06 &0.20 \\ \hline
    $\bar\omega$& 0.27 & 0.55 &0.27 &0.46 &0.23 &0.09 &0.12 &0.38 \\ \hline
   $\Omega$ & 0.22 & 0.55 & 0.23 & 0.34 & 0.18 & 0.08 & 0.08 & 0.32 \\ \hline
    $\Delta$ & 0.09 & 0 & 0.09 & 0.20 & 0.07 & 0.04 & 0.05 & 0.18 \\ \hline
    $\sigma$& 0.05 & 0 & 0.05 & 0.06 & 0.02 & 0.01 & 0.02 & 0.06 \\ \hline
    $\sigma$\%& 22.00 & 0.01 & 20.78 & 18.83 & 12.11 & 15.82 & 18.90 & 18.02 \\ \hline
    T & 0.27 & 0.27 & 0.27 & 0.2 & 0.2 & 0.2 & 0.2 & 0.2 \\ \hline
    H1:$\Delta$\textgreater T & OK & OK & OK & OK & OK & OK & OK & OK \\ \hline
    \end{tabular}} \\ 
      \footnotesize{$\underline{\omega}$: Lower bound of weight, $\bar\omega$: Upper bound of weight , $\Omega$: Point estimate of weight (OPA $\omega$), $\Delta$: Length of the interval , $\sigma$: Standard deviation, $\sigma$ \%: Coefficient of variation , T: Threshold value.}
    \end{table}

Finally, sensitivity analysis was performed to assess the reliability of the result. The results suggest that the results are robust, and rankings of the factors do not change because of small variations in the decision maker's input. The visual representation of sensitivity analysis is depicted in Fig. 7. 

\begin{table*}
\label{fig_7}
\begin{tabular}{p{0.49\textwidth}p{0.49\textwidth}}
   \begin{tikzpicture}
\begin{axis}[
ylabel={\scriptsize Global Weight},
xmin=1, xmax=5,
ymin=0.0, ymax=0.6,
xtick={1,2,3,4,5},xticklabels={$F_1$,$F_2$,$F_3$,$F_4$,$F_5$},
xticklabel style = {font=\scriptsize},
ytick={0,0.15,0.30,0.45,0.60},
yticklabel style = {font=\scriptsize},
legend pos=north east,
legend style={font=\scriptsize},
ymajorgrids=true,xmajorgrids=true,
grid style=dashed,
height=6cm,
width=3.2in
]

\addplot[color=teal,mark=square, very thick]
coordinates {(1,0.45)(2,0.23)(3,0.09)(4,0.11)(5,0.38)};
\addlegendentry{Lower $\omega$}
\addplot[color=blue,mark=triangle, very thick]
coordinates {(1,0.345)(2,0.19)(3,0.08)(4,0.095)(5,0.32)};
\addlegendentry{Upper $\omega$}
\addplot[color=brown,mark=otimes, very thick]
coordinates {(1,0.245)(2,0.155)(3,0.06)(4,0.075)(5,0.2)};
\addlegendentry{OPA $\omega$}

\end{axis} \label{fig_first_case}
\end{tikzpicture} 
 & 
\begin{tikzpicture}
\begin{axis}[
ylabel={\scriptsize Global Weight},
xmin=1, xmax=5,
ymin=0.0, ymax=0.6,
xtick={1,2,3,4,5},xticklabels={$F_1$,$F_2$,$F_3$,$F_4$,$F_5$},
xticklabel style = {font=\scriptsize},
ytick={0,0.15,0.30,0.45,0.60},
yticklabel style = {font=\scriptsize},
legend pos=north east,
legend style={font=\scriptsize},
ymajorgrids=true,xmajorgrids=true,
grid style=dashed,
height=6cm,
width=3.2in
]

\addplot[color=teal,mark=square, very thick]
coordinates {(1,0.45)(2,0.23)(3,0.09)(4,0.11)(5,0.38)};
\addlegendentry{Lower $\omega$}
\addplot[color=blue,mark=triangle, very thick]
coordinates {(1,0.345)(2,0.19)(3,0.08)(4,0.095)(5,0.32)};
\addlegendentry{Upper $\omega$}
\addplot[color=brown,mark=otimes, very thick]
coordinates {(1,0.245)(2,0.155)(3,0.06)(4,0.075)(5,0.2)};
\addlegendentry{OPA $\omega$}

\end{axis} \label{fig_second_case}
\end{tikzpicture}\\
\centering{(a) Criteria} & \centering{(b) Factors} \\
\end{tabular}
Fig. 7. Sensitivity Analysis of the Results. 
      
\end{table*}
\subsection{Discussion}
The results of this study suggest that different standards have different key concepts. Some of these concepts, such as the right to consent, are repeated in each standard \cite{c17article}. Extant literature reflects that none of these standards is sufficient to implement privacy and security in the context of EHRs. Moreover, all these standards are taken from developed countries and are not wide-ranging. Hence, the study selected popular standards from the USA and Europe and the newly passed DPDPA act from India to make the selection more inclusive. The prioritization of efforts is key in the case of resource-constrained settings such as developing countries; hence, the study first grouped twenty identified key factors into five factors and ranked these factors. The study concluded that the efficacy of implementing the factors is most important, followed by cost and ease of implementation. The study concluded that technical rights are the most important factor for implementing privacy and security. The significant finding of the study is that operational practices play a key role in successful implementation.  

The integration of EHRs into healthcare systems has revolutionized patient care, offering seamless access to patient history, diagnostic results, and treatment plans. However, this digitization comes with significant privacy and security concerns. According to a study by \cite{JIANG2018392}, the main issues in EHR security arise from unauthorized access, data breaches, and the potential for data manipulation. This can lead to loss of patient trust and legal complications. Additionally, \cite{cyber927398} highlight the risks of cyber-attacks, which have become more sophisticated, making EHR systems vulnerable. This research highlights the key concepts related to data privacy and security and provides a comprehensive framework for its implementation. 

Consumer electronics, such as smartphones and tablets, have become pivotal in accessing EHRs, especially with the growth of telemedicine. According to \cite{mohamed_chawki_2021_4782466} mobile devices enable healthcare providers and patients to access health records remotely, enhancing the efficiency of healthcare delivery. However, this also opens new avenues for potential security breaches. \cite{Chernyshev2018} argues that while consumer electronics make EHRs more accessible, they also pose significant security risks due to their susceptibility to loss, theft, and hacking. The use of personal devices for accessing EHRs demands stringent security protocols, including secure VPNs, regular software updates, and strong password policies.

HIPAA in the United States sets a standard for protecting sensitive patient data. Failure to comply with these standards can result in severe penalties, as discussed by \cite{KHAN2018395}. Furthermore, ethical considerations in managing EHRs, as noted by \cite{GOODMAN201658}, include ensuring patient confidentiality and informed consent for data usage. As EHRs become integral in clinical research and public health monitoring, maintaining ethical standards in data usage is paramount. EHRs have significantly improved healthcare delivery, they present notable challenges in terms of data privacy and security. The role of consumer electronics in accessing these records introduces additional complexities that require careful management. Adhering to regulatory standards, employing advanced security measures, and considering ethical implications are essential for maintaining the integrity of EHR systems.

\section{Conclusion} \label{seccon}
Recent years have seen significant digitalization in healthcare, accelerated by the pandemic and technological advancements. Despite initial hesitancy, the healthcare sector is now rapidly adopting technology, particularly in the digitalization of Electronic Health Records (EHRs), considered the backbone of digital health. This study reviews five EHR standards, identifying twenty key concepts through expert evaluations. Utilizing National Institute of Standards and Technology dimensions, K-means clustering identifies five crucial factors: Technical Rights, Admin Rights, Security Measures, Data Protection, and Operational Practices. Efficacy proves the most critical criterion, followed by cost and ease of implementation, with Technical Rights and Operational Practices holding the highest significance. The study contributes both theoretically and practically by providing a comprehensive account of key concepts associated with privacy and security standards, along with a prescriptive framework for EHR implementation. Future work will focus on user-centric evaluations to enhance standards' adaptability to evolving electronic healthcare data management needs.



\bibliographystyle{IEEEtran}
\bibliography{reference}

\begin{IEEEbiography}[{\includegraphics[width=1in,height=1.25in,clip,keepaspectratio]{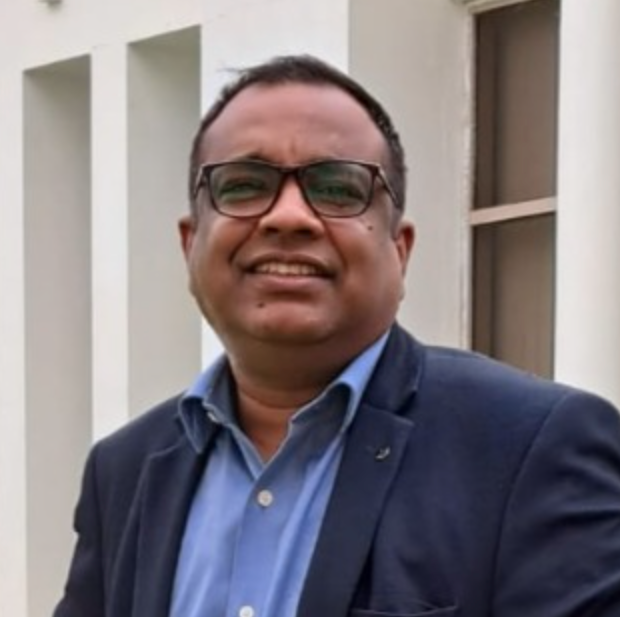}}]{Vinaytosh Mishra}received the Bachelor of Technology degree in electronics engineering and the Ph.D. degree in healthcare management from the Indian Institute of Technology (BHU), India. He was a Postdoctoral Fellow of AI in Healthcare with the University of Arizona, USA, and a Postdoctoral Fellow of Ethical AI in Healthcare with the University of Ben Gurion, Israel. He is an Associate Professor and the Director of the Thumbay Institute for AI in Healthcare, Gulf Medical University, Ajman, UAE. He has over 19 years of experience in industries, such as information technology, manufacturing, finance, healthcare, and education alongwith one Australian and one German Patent in AI in Healthcare. He has published more than 70 research papers in different journals and conferences of high repute. His current research interests include statistics, quality assurance engineering, supply chain management, and healthcare digital health. 
\end{IEEEbiography}
\vspace{11pt}
\begin{IEEEbiography}
[{\includegraphics[width=1in,height=1.25in,clip,keepaspectratio]{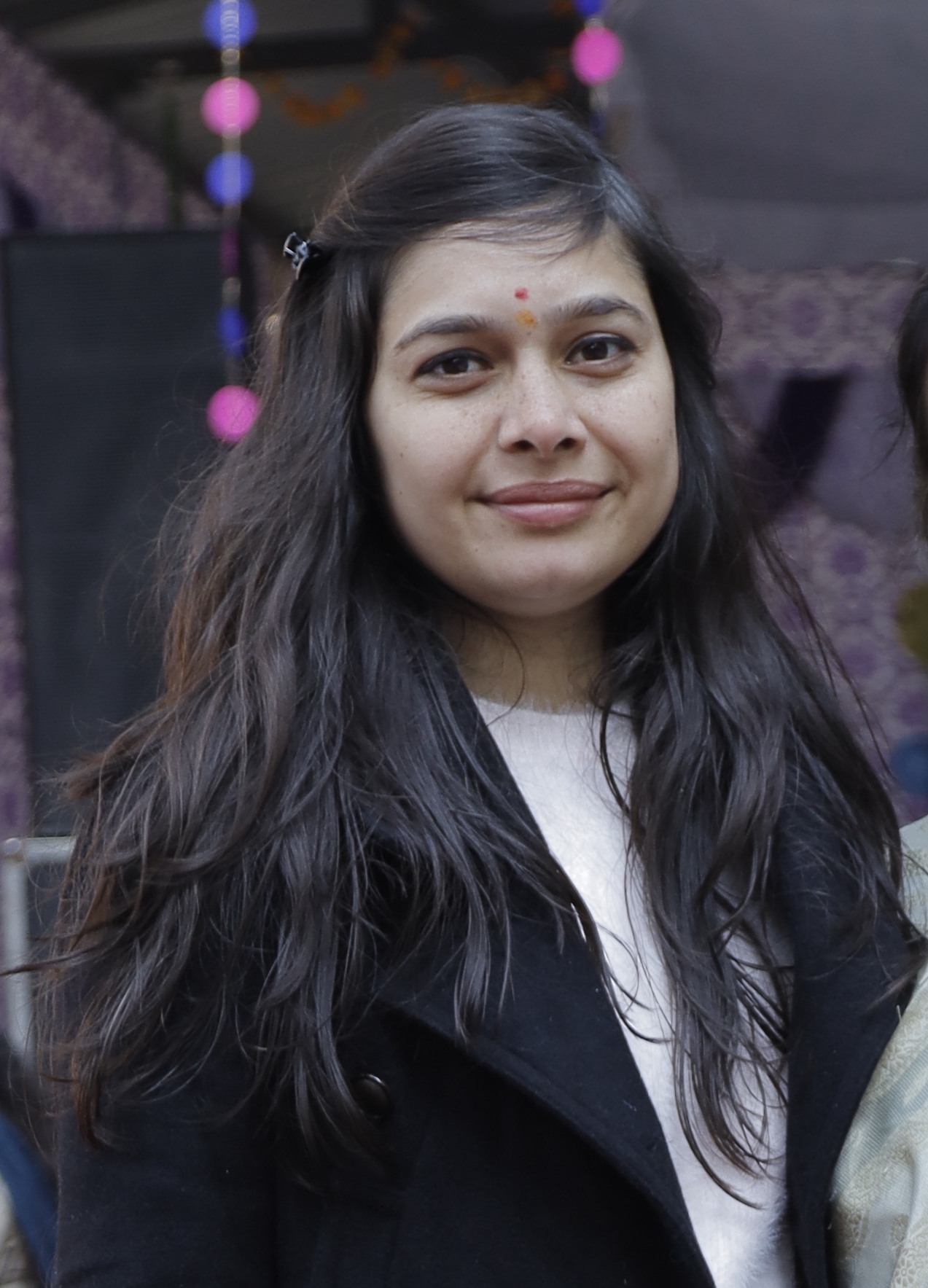}}]{Kishu Gupta}(Member, IEEE) received the M.Sc. degree in computer science, and the M.C.A. and Ph.D. degrees from Kurukshetra University,
Kurukshetra, India, in 2014, 2015, and 2023, respectively. She is working as a Post Doctoral Research Fellow with the Cloud Computing Research Center, Department of Computer Science and Engineering, National Sun Yat-sen University, Kaohsiung, Taiwan. Her research interest includes data security and privacy, cloud computing, federated learning, machine learning, and quantum computing. She is a recipient of the Gold Medal for securing Ist rank in overall university during M.Sc. Also, she has been awarded with prestigious INSPIRE Fellowship sponsored by the Department of Science and Technology, under the Ministry of Science and Technology, Government of India.
\end{IEEEbiography}
\vspace{11pt}
\begin{IEEEbiography}[{\includegraphics[width=1in,height=1.25in,clip,keepaspectratio]{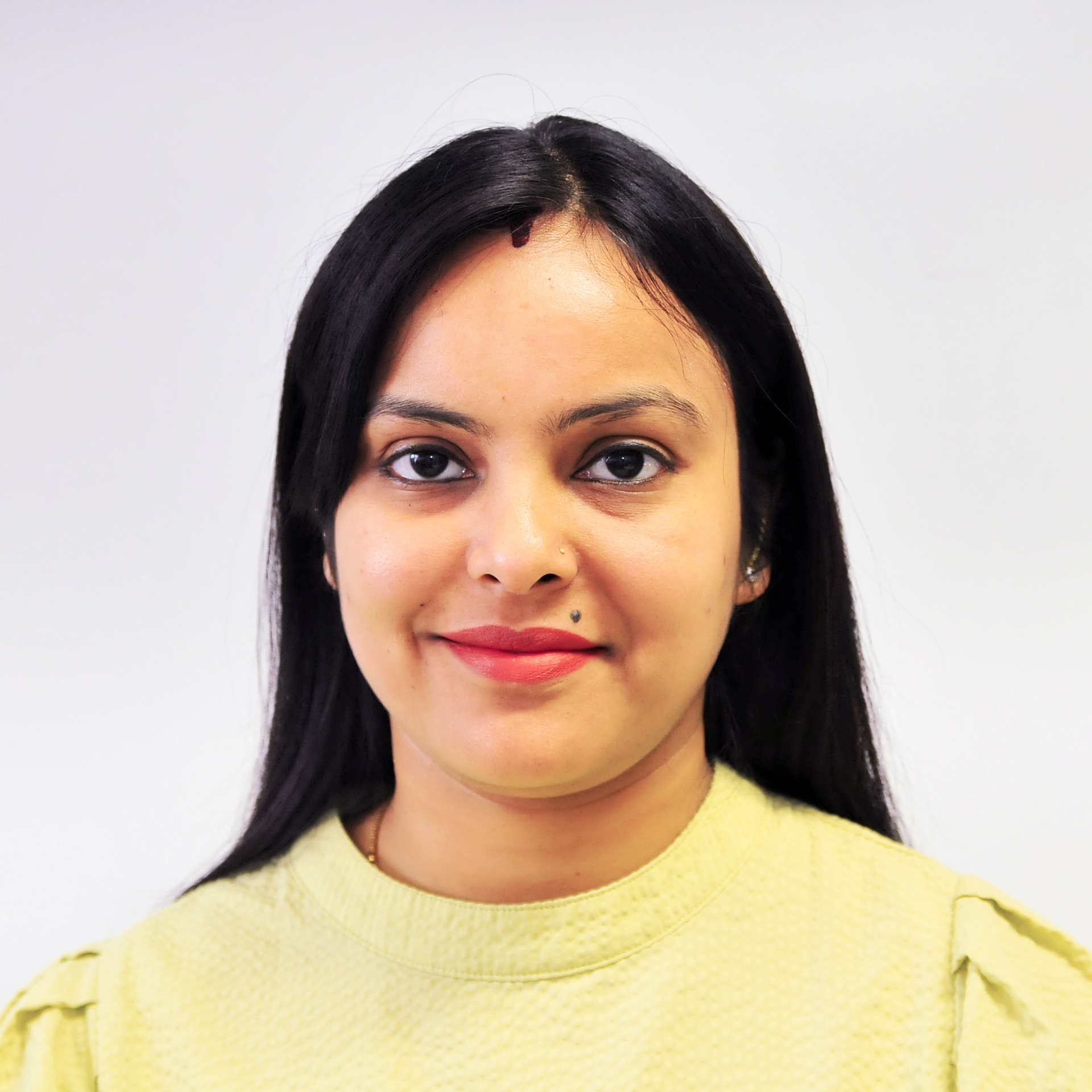}}]{Deepika Saxena}(Member, IEEE) received the Ph.D. degree in computer science from the National Institute of Technology, Kurukshetra, India, and the Postdoctoral degree from the Department of Computer Science, Goethe University, Frankfurt, Germany. She holds the position of an Associate Professor with the Division of Information Systems, The University of Aizu, Japan. Also, she is working as a part-time Faculty with the University of Economics and Human Sciences, Warsaw, Poland, Europe. Some of her research findings are published in top cited journals, such as IEEE TRANSACTIONS ON AUTOMATION SCIENCE AND ENGINEERING, IEEE TRANSACTIONS ON SERVICES COMPUTING, IEEE TRANSACTIONS ON SYSTEMS, MAN, AND CYBERNETICS, IEEE TRANSACTIONS ON GREEN COMMUNICATIONS AND NETWORKING, IEEE TRANSACTIONS ON PARALLEL AND DISTRIBUTED SYSTEMS, IEEE TRANSACTIONS ON CLOUD COMPUTING, IEEE COMMUNICATIONS LETTERS, IEEE NETWORKING LETTERS, IEEE SYSTEMS JOURNAL, IEEE WIRELESS COMMUNICATION LETTERS, IEEE TRANSACTIONS ON NETWORK AND SERVICE MANAGEMENT, IET Electronics Letters, and Neurocomputing. Her major research interests include neural networks, evolutionary algorithms, resource management and security in cloud computing, internet traffic management, and quantum machine learning, datalakes, and dynamic caching management. She is the recipient of the prestigious IEEE TCSC 2023 Outstanding Ph.D. Dissertation Award and the EUROSIM 2023 Best Ph.D. Thesis Award. Also, her research paper, published in the IEEE TRANSACTIONS ON CLOUD COMPUTING, received the 2022 Best Paper Award from the IEEE Computer Society Publications Board.
\end{IEEEbiography}
\vspace{11pt}
\begin{IEEEbiography}[{\includegraphics[width=1in,height=1.25in,clip,keepaspectratio]{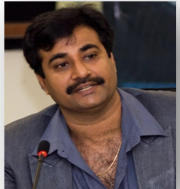}}]{Ashutosh Kumar Singh} (Senior Member, IEEE) received the Ph.D. degree in electronics engineering from the Indian Institute of Technology, BHU, India, and the Postdoctoral degree from the Department of Computer Science, University of Bristol, U.K. He is working as a Professor and the Director of the Indian Institute of Information Technology Bhopal, India. Also, he is working as an Adjunct Professor with the University of Economics and Human Sciences, Warsaw, Poland. He has research and teaching experience in various Universities of the India, U.K., and Malaysia. He has published more than 370 research papers in different journals and conferences of high repute. Some of his research findings are published in top cited journals, such as IEEE TRANSACTIONS ON SERVICES COMPUTING, IEEE TRANSACTIONS ON COMPUTERS, IEEE TRANSACTIONS ON SYSTEMS, MAN, AND CYBERNETICS, IEEE TRANSACTIONS ON PARALLEL AND DISTRIBUTED SYSTEMS, IEEE TRANSACTIONS ON INDUSTRIAL INFORMATICS, IEEE TRANSACTIONS ON CLOUD COMPUTING, IEEE COMMUNICATIONS LETTERS, IEEE NETWORKING LETTERS, IEEE Design \& Test, IEEE SYSTEMS JOURNAL, IEEE WIRELESS COMMUNICATION LETTERS, IEEE TRANSACTIONS ON NETWORK AND SERVICE MANAGEMENT, IEEE TRANSACTIONS ON GREEN COMMUNICATIONS AND NETWORKING, IET Electronics Letters, Future Generation Computer Systems, Neurocomputing, Information Sciences, and Information Processing Letters. His research area includes design and testing of digital circuits, data science, cloud computing, machine learning, and security. His research paper, published in the IEEE TRANSACTIONS ON CLOUD COMPUTING, was honored with the 2022 Best Paper Award by the IEEE Computer Society Publications Board.
\end{IEEEbiography}

\vfill

\end{document}